%% file: ms.tex
\documentclass[sigconf]{acmart}

\AtBeginDocument{%
  \providecommand\BibTeX{{%
    \normalfont B\kern-0.5em{\scshape i\kern-0.25em b}\kern-0.8em\TeX}}}

\usepackage{subcaption}
\usepackage{graphicx}
\usepackage{amsthm}
\usepackage{multirow}
\usepackage{rotating}
\usepackage{framed}
\newtheorem{definition}{Definition}[section]

\newcommand{\nummodel}{$14$}
\newcommand{\ccnum}{$8$\space}
\newcommand{\metricnum}{$25$\space}
\newcommand{\nn}{\mathcal{D}}

\usepackage{xcolor}

\usepackage[normalem]{ulem}

\begin{document}

\title{There is Limited Correlation between Coverage and Robustness for Deep Neural Networks}
\title{There is Limited Correlation between Coverage and Robustness for Deep Neural Networks}
\author{Yizhen Dong}
\authornote{Both authors contributed equally to this research.}
\email{dyz386846383@163.com}
\affiliation{%
  \institution{Tianjin University}
  \country{China}
}
\author{Peixin Zhang}
\authornotemark[1]
\email{pxzhang94@zju.edu.cn}
\affiliation{%
  \institution{Zhejiang University}
  \country{China}
}
\author{Jingyi Wang}
\affiliation{%
  \institution{National University of Singapore}}
\email{wangjy@comp.nus.edu.sg}
\author{Shuang Liu}
\affiliation{%
  \institution{Tianjin University}}
\email{shuang.liu@tju.edu.cn}
\author{Jun Sun}
\affiliation{%
  \institution{Singapore Management University}}
\email{junsun@smu.edu.sg}
\author{Ting Dai}
\affiliation{%
  \institution{Huawei Corporate}}
\email{daiting2@huawei.com}
\author{Xinyu Wang}
\affiliation{%
  \institution{Zhejiang University}}
\email{wangxinyu@zju.edu.cn}
\author{Jianye Hao}
\affiliation{%
  \institution{Tianjin University}}
\email{jianye.hao@tju.edu.cn}
\author{Li Wang}
\affiliation{%
  \institution{Tianjin University}}
\email{wangli@tju.edu.cn}
\author{Jin Song Dong}
\affiliation{%
  \institution{National University of Singapore}}
\email{dongjs@comp.nus.edu.sg}

\date{}

\thispagestyle{empty}

\begin{abstract}

Deep neural networks (DNN) are increasingly applied in safety-critical systems, e.g., for face recognition, autonomous car control and malware detection. It is also shown that DNNs are subject to attacks such as adversarial perturbation and thus must be properly tested. Many coverage criteria for DNN since have been proposed, inspired by the success of code coverage criteria for software programs. The expectation is that if a DNN is a well tested (and retrained) according to such coverage criteria, it is more likely to be robust.
In this work, we conduct an empirical study to evaluate the relationship between coverage, robustness and attack/defense metrics for DNN. Our study is the largest to date and systematically done based on $100$ DNN models and \metricnum metrics. One of our findings is that there is limited correlation between coverage and robustness, i.e., improving coverage does not help improve the robustness. Our dataset and implementation have been made available to serve as a benchmark for future studies on testing DNN.

\end{abstract}
\maketitle
\input{1_Introduction}

\input{3_Preliminary}

\input{4_Experiment}

\input{4_Implementation}
\input{5_Discussion}
\input{2_Related_Work}
\input{6_Conclusion}
\clearpage



\bibliographystyle{plainnat}
\bibliography{ms}

\end{document}

%% file: 1_Introduction.tex
\section{Introduction}
\label{sec:introduction}
Recent years have seen rapid development on deep learning techniques as well as applications in a variety of domains like computer vision~\cite{ResNet,vgg} and natural language processing~\cite{bert}. There is a growing trend to apply deep learning for solving safety-critical tasks, such as face recognition~\cite{face_recognition}, self-driving cars~\cite{selfdriving} and malware detection~\cite{malware}. Unfortunately, deep neural networks (DNN) are shown to be vulnerable to attacks and lack of robustness. For instance, they are easily subject to adversarial perturbation~\cite{cw, fgsm}, i.e., a DNN makes a wrong decision given a carefully crafted small perturbation on the original input. Such attacks have been demonstrated successfully in the physical world~\cite{DBLP:journals/corr/KurakinGB16}. This suggests that DNN, just like software systems, must be properly analyzed and tested before they are applied in safety-critical systems.

The software engineering community welcomed the challenge and opportunity. Multiple software testing approaches, i.e., differential testing~\cite{deepxplore}, mutation testing~\cite{DeepMutation,ouricse19} and concolic testing~\cite{DeepConcolic}, have been adapted into the context of testing DNN. Inspired by the noticeable success of code coverage criteria in testing traditional software systems, multiple coverage criteria\footnote{Metric and criterion are used interchangeably.}, e.g., neuron coverage~\cite{deepxplore, DeepTest} and its extensions DeepGauge~\cite{DeepGauge}, MC/DC~\cite{DeepCover}, and Surprise Adequacy~\cite{SurpriseAdequacy}, have been proposed. Coverage criteria quantitatively measures how well a DNN is tested and offers guidelines on how to create new test cases. The underlying assumption is that a DNN which is better tested, i.e., with higher coverage, is more likely to be robust.

This assumption however is often not examined or only evaluated with limited DNN models and structures, making it unclear whether the results generalize. Furthermore, how a test suite improves the quality of a DNN is different from that of a software system. A software system is improved by fixing bugs revealed by a test suite. A DNN is typically improved by retraining with the test suite. While existing studies show that retraining often improves a DNN's accuracy to some extent~\cite{deepxplore, DeepConcolic}, it is not clear whether there is correlation between the coverage of the test suite and the improvement, i.e., does a set of inputs with higher coverage imply better improvement (on DNN robustness)?



Inspired by the work in~\cite{DBLP:conf/icse/InozemtsevaH14}, we conduct an empirical study to evaluate whether coverage is correlated with robustness of DNN and additional metrics which are associated with the quality of DNN~\cite{DeepSec}. In particular, we would like to answer the following research questions.
\begin{itemize}
\item Are there correlations between testing coverage criteria and the robustness of DNN?
\item Are there correlations among different coverage criteria themselves?
\item Are there correlations between the improvement of coverage criteria and the improvement in terms of robustness after the DNN is retrained?
\item Are there metrics that are strongly correlated to the robustness of DNN or the robustness improvement after retraining?
\end{itemize}
Based on the answers to the above questions, we aim to provide practical guidelines for developing testing methods which contribute towards improving the robustness of DNN.

Conducting such an empirical study is highly non-trivial. First, we need a large set of real-world DNN for the study. However, training realistic DNN often takes significant amount of time and resource. For instance, it takes $15$ GPU hours to train a ResNet-101 model.
Our study trained $100$ state-of-the-art DNN models\footnote{$25$ seed models trained with original dataset and $75$ models retrained using  original dataset augmented with adversarial samples.} with a variety of architectures with two popular datasets, i.e., MNIST~\cite{MNIST_Data} and CIFAR10~\cite{Cifar}. Obtaining these models took a total of 150 GPU hours.

Second, we need to obtain adversarial samples by attacking the trained original models. We adopt $3$ state-of-the-art attack methods, i.e., FGSM~\cite{fgsm}, JSMA~\cite{jsma} and C\&W~\cite{cw}, to attack the original models, in order to obtain different adversarial sample sets and train different DNN models. Some of the adversarial attack methods, e.g., JSMA and C\&W, are known to be time-consuming. It takes us a total of $1,810$ GPU hours to obtain adversarial samples for all the original models with the $3$ attack methods.

Last but not least, we need a systematic and automatic way of evaluating the coverage, robustness, and other associated metrics, which is not always straightforward. For instance, there are multiple definitions of robustness in the literature~\cite{Robustness},~\cite{Clever},
some of which are complicated and expensive to compute (e.g., it took $12$ GPU hours to compute a CLEVER score~\cite{Clever} for GoogLeNet-22.). In this work, we develop a self-contained toolkit called \emph{DRTest} (\emph{D}eep \emph{R}obustness \emph{T}esting), which calculates a comprehensive set of metrics on DNN, including 1) \ccnum testing coverage criteria proposed for DNN, 2) $2$ robustness metrics for DNN, and 3) a set of $15$ attack and defense metrics for DNN. A total of $4,150$ GPU hours are spent on computing these metrics based on the above-mentioned models.

Our empirical study is conducted as follows. For each dataset, we first train $25$ diverse seed models (with state-of-the-art architectures), attack each seed model with different attacking methods to generate adversarial samples (with varying attack parameters), augment the training dataset with the generated adversarial samples, and retrain the model. We apply \emph{DRTest} to calculate a range of metrics for every model. Afterwards, we apply a standard correlation analysis algorithm, the  Kendall's rank correlation coefficient~\cite{Kendall}, to analyze the correlations between the metrics. 




In summary, we make the following contributions.
\begin{itemize}
	\item We conducted an empirical study to systematically investigate the correlation between coverage, robustness and related metrics for DNN. 
Based on the empirical study results, we discuss potential research directions on DNN testing.
	\item We implemented a self-contained and extensible toolkit which calculates a large set of metrics, which can be used to quantitatively measure different aspects of DNN.
	\item We publish online our models, adversarial samples, retrained models as well as \emph{DRTest}, which can be used as a benchmark for future proposals on methods for DNN testing.
\end{itemize}

We organize the remainder of the paper as follows. Section~\ref{sec:pre} introduces the background knowledge of this work. Section~\ref{sec:meth} presents our research methodology. Section~\ref{sec:implementation} shows details on our implementations. Section~\ref{sec:exp} reports our findings on the research questions. We present related works in Section~\ref{sec:related} and conclude in Section~\ref{sec:con}.


%% file: 3_Preliminary.tex
\section{Preliminaries}
\label{sec:pre}
In this section, we briefly review preliminaries related to this work, which include Deep Neural Networks (DNN), adversarial attacks on DNN, testing methods for DNN, and robustness of DNN. 



%
%
\subsection{Deep Neural Networks}
\label{subsec:DNN}


A DNN is an artificial neural network with multiple layers between the input and output layers. It can be denoted as a tuple $\mathcal{D}=(\mathnormal{I}, \mathnormal{L}, \mathnormal{F}, \mathnormal{T})$ where
\begin{itemize}
\item$\mathnormal{I}$ is the input layer;
\item$\mathnormal{L} = \{ \mathnormal{L}_{\mathnormal{j}}|\mathnormal{j}\in\{1, \dots, \mathnormal{J}\}\}$ is a set of hidden layers and the output layer, each of which contains $\mathnormal{s}_{\mathnormal{j}}$ neurons, and the $\mathnormal{k}_{th}$ neuron in layer $\mathnormal{L}_{\mathnormal{j}}$ is  denoted as $\mathnormal{n}_{{\mathnormal{j},\mathnormal{k}}}$ and its value is $\mathnormal{v}_{{\mathnormal{j},\mathnormal{k}}}$;
\item$\mathnormal{F}$ is a set of activation functions;    
\item$\mathnormal{T}$: $\mathnormal{L} \times \mathnormal{F} \to \mathnormal{L}$ is a set of transitions between layers. The output of each neuron is computed by applying the activation function to the weighted sum of its inputs, and the weights represent  the strength of the connections between two linked neurons.
\end{itemize}

In this work, we focus on DNN classifiers $\mathcal{D}(X):X \to Y$, where $X$ is a set of inputs and $Y$ is a finite set of labels. Given an input $x\in X$, a DNN classifier transforms information layer by layer and outputs a label $y\in Y$ for the input $x$. In this work, we try to cover a wide range of (including state-of-the-art) DNN architectures. We briefly introduce them in the following.

\begin{figure}[tp]
\includegraphics[scale=0.3]{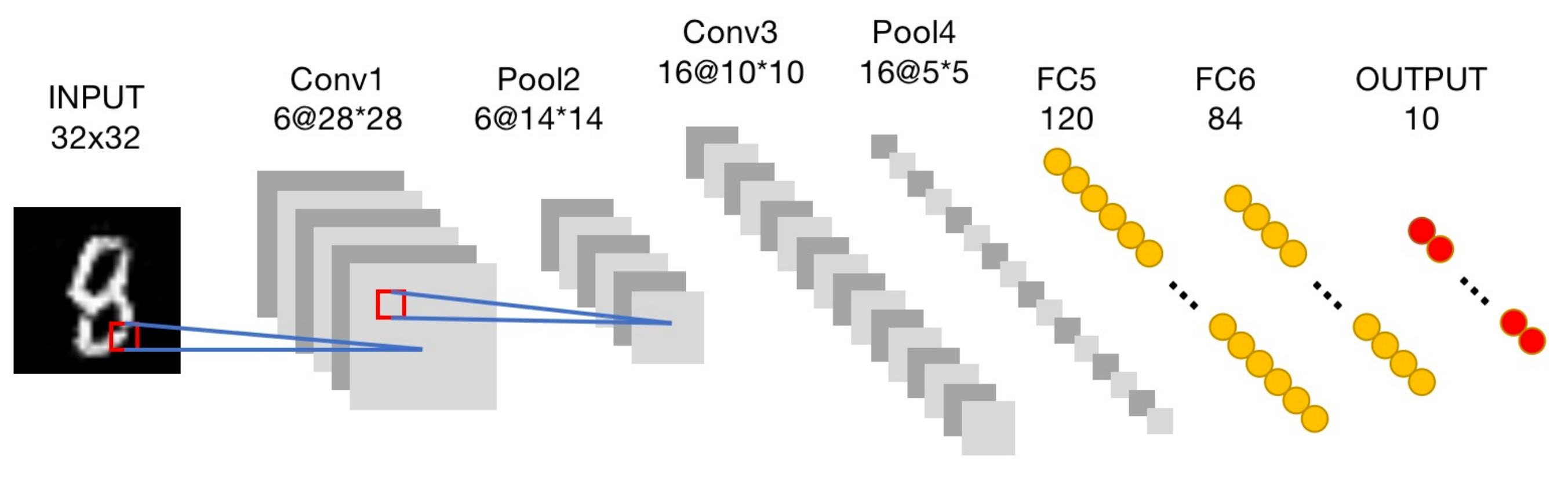}
\centering
\caption{LeNet-5 Structure}
\label{fig:LeNet-5}
\vspace{-4mm}
\end{figure}

\vspace{1mm}
\noindent\textbf{LeNet}~\cite{lenet} is one of the most representative DNN architectures. As shown in Fig.~\ref{fig:LeNet-5}, the basic modules include the convolution layers (Conv), the pooling layers (Pool) and the fully connected layers (FC). Conv aims to extract different local features and Pool makes sure to get the same feature after transformation, i.e., translation, rotation, and scaling. FC then maps the distributed feature representations from Conv and Pool to the label space.

\vspace{1mm}
\noindent\textbf{VGG}~\cite{vgg} is an advanced architecture for extracting CNN features from images. Compared to LeNet, VGG utilizes smaller convolution kernels (e.g., $3*3$ or $1*1$) and pooling kernels (e.g., $2*2$) which significantly increases the expressive power.

\vspace{1mm}
\noindent\textbf{GoogLeNet}~\cite{googlenet} Unlike most popular DNN models which obtain better accuracy by increasing the depth of the network, it introduces an inception module with a parallel topology to expand the width of the model instead. The inception module helps to extract richer features and reduce dimensions using $1*1$ convolution kernel, and aggregate convolution results on multiple sizes to obtain features from different scales and accelerate convergence rate.

\vspace{1mm}
\noindent\textbf{ResNet}~\cite{ResNet} improves traditional sequential CNNs by solving the vanishing gradients problem when expanding the number of layers. It utilizes short-cuts (also called skip connections), which adds up the input and output of a layer and then transforms the sum to the next layer as input.

\subsection{Adversarial Attack}
\label{sec:attackmethod}
Since Szegedy \emph{et al.} discovered that DNNs are intrinsically vulnerable to adversarial samples (i.e., sample inputs which are generated through perturbation with the intention to trick a DNN into wrong decisions)~\cite{Intriguing}, many attacking approaches have been developed to craft adversarial samples. In the following, we briefly introduce $3$ popular attacking algorithms that we adopt in our work.

\vspace{1mm}
\noindent\textbf{FGSM} Goodfellow \emph{et al.} proposed the first and fastest attacking algorithm Fast Gradient Sign Method (FGSM)~\cite{fgsm}, which attempts to maximize the change of probability of sample's original label by the gradient of its loss. The implementation of FGSM is as follows, which is quite straightforward and efficient.
\begin{equation}
x'=x+\epsilon \cdot \textbf{sign}(\nabla_{x} J(x, c_{x}))
\end{equation}
$J$ is the loss function for training, $c_{x}$ is the prediction of x and $\epsilon$ is a hyper-parameter to control the degree of perturbation. 

\vspace{1mm}
\noindent\textbf{JSMA} Jacobian-based Saliency Map Attack (JSMA)~\cite{jsma} is a targeted attack method. First, it calculates a saliency map based on the Jacobian matrix of a given sample. Each value of the map represents the impact of the corresponding pixel to the target prediction. Then it greedily picks the most influential features each time and maximizes their values until either successfully generates an adversarial sample or the number of pixels modified exceeds the bound. We refer the readers to~\cite{jsma} for details.

\vspace{1mm}
\noindent\textbf{C\&W} Carlini \emph{et al.}~\cite{cw} aim to craft adversarial samples with high confidence and small perturbation based on certain distance metric by solving the following optimization problem directly:
\begin{equation}
\arg \min ||x'-x||_{p} + \lambda \cdot f(x',t)
\end{equation}
$||x'-x||_{p}$ is the perturbation according to p-norm measurements, e.g., $L_{0}$, $L_{2}$ and $L_{\infty}$; $t$ is the target label and $\lambda$ is a hyper-parameter to balance the objectives. In order to prevent adversarial samples from generating illegal values, they devised a group of clip functions and loss functions. Readers can refer to~\cite{cw} for details.

\subsection{Testing Deep Neural Networks}

A variety of traditional software testing methods like differential testing~\cite{differential1, differential2}, concolic testing~\cite{MCDC} have been adapted to the context of testing DNN~\cite{deepxplore, DeepConcolic} to find adversarial samples (in hope of revealing bugs in DNN). Note that in the setting of DNN testing, a test case is a sample input. In the following, we review some recently proposed coverage criteria for DNN.

\vspace{1mm}
\noindent\textbf{Neuron Coverage} Neuron coverage~\cite{deepxplore} is the first coverage criteria proposed for testing DNN, which quantifies the percentage of activated neurons by at least one test case in the test suite. The authors also proposed a differential testing method to generate test cases to improve neuron coverage. 

\vspace{1mm}
\noindent\textbf{DeepGauge} Later, Ma \emph{et al.} proposed DeepGauge~\cite{DeepGauge}, which extends neuron coverage with coverage criteria which are defined based on the activation values from two different levels. For instance, \emph{neuron-level} coverage first divides the range of values at each neuron into $k$ sections during the training stage, obtains the upper and lower bounds, and then evaluates if each section is covered or the boundary has been crossed by the test suite. The \emph{layer-level} coverage concerns how many neurons used to be the top-$k$ active neurons are activated at least once for each layer (TKNC), or whether the pattern formed by sequences of top-$k$ active neurons on each layer (TKNP) is present. 

\vspace{1mm}
\noindent\textbf{Surprise Adequacy} Based on the idea that a good test  suite should be `surprising' compared to the training set, Kim \emph{et al.}~\cite{SurpriseAdequacy} defined two measures on how surprising a testing input is to the training set. One is called the kernel density estimation, which evaluates the likelihood of the testing input from the training set. The other measures the Euclidean distance of neuron activation traces for a given input and the training set. Readers are referred to~\cite{SurpriseAdequacy} for details.




\subsection{Robustness of Deep Neural Networks}
Given the existence of adversarial samples, adversarial robustness becomes an important desired property of a DNN which measures its resilience against adversarial perturbations. Following the definitions proposed by Katz et al.~\cite{katz2017reluplex}, adversarial robustness can be categorized into local adversarial robustness and global adversarial robustness depending on different contexts.

\begin{definition}
(Local Adversarial Robustness) Given a sample input $x$, a DNN $\nn$ and a perturbation threshold $\epsilon$, $\nn$ is $\epsilon-$local robust iff for any sample input $x'$ such that $||x-x'||_p\le\delta$, we have $\nn(x)=\nn(x')$, where $||\cdot||_p$ is the $p$-norm to measure the distance between two sample inputs.
\end{definition}  

\begin{definition}
(Global Adversarial Robustness) For any sample inputs $x$ and $x'$, a DNN $\nn$ and two thresholds $\delta,\epsilon$, $\nn$ is $(\delta,\epsilon)-$robust iff for any $||x-x'||_p\le\delta$, we have $|\nn(x)-\nn(x')|\le\epsilon$.  
\end{definition}

Local robustness measures the robustness on a specific input, while global robustness measures the robustness on all inputs. 

Verifying whether a DNN satisfies local or global robustness is an active research area~\cite{katz2017reluplex,AI2,Formalguarantees} and existing methods do not scale to state-of-the-art DNNs (especially for global robustness). Thus, multiple metrics have been proposed in order to empirically evaluate the adversarial robustness of a DNN in the literature~\cite{Clever,Robustness,virmaux2018lipschitz,fazlyab2019efficient}. In the following, we introduce two widely used adversarial robustness metrics (including both local~\cite{Clever} and global~\cite{Robustness} robustness) which we adopt in this work.


\vspace{1mm}
\noindent\textbf{Global Lipschitz Constant}
Lipschitz Constant~\cite{Robustness} measures the sensitivity of a model to adversarial samples. Given a function $\mathnormal{f}$, its Lipschitz constant is only related to the parameters of $\mathnormal{f}$. In our context, the function is in the form of a DNN. Its Lipschitz constant can be calculated recursively layer-by-layer from the output layer all the way to the input layer,  taking consideration of short-cuts in ResNet and inception module in GoogLeNet.
For example, the Lipschitz Constant of a DNN which has a structure similar to LeNet and VGG is the product of the Lipschitz Constant of all the hidden layers and the output layer. 

As an example, we introduce how Lipschitz Constant is calculated for a fully connected layer. Readers are referred to~\cite{Parseval} for the calculation of convolution and aggregation layers. 
Let $ \mathbf{v}_{j-1}$ and $\hat{\mathbf {v}}_{j-1}$ be two inputs of layer $\mathnormal{L}_{\mathnormal{j}}$; $ \mathbf{v}_{j}$ and $\hat{\mathbf {v}}_{j}$ be their corresponding outputs; and $\omega_{i,k}^{j}$ be the parameter of the connection between the $\mathnormal{k}_{th}$ neuron in layer $\mathnormal{L}_{\mathnormal{j}}$  and the $\mathnormal{i}_{th}$ neuron in layer $\mathnormal{L}_{\mathnormal{j-1}}$; and $\mathnormal{s}_{\mathnormal{j}}$ be the number of neurons of layer $\mathnormal{L}_{\mathnormal{j}}$. The Lipschitz Constant for layer $\mathnormal{L}_{\mathnormal{j}}$ is defined as $\alpha = \max _ { k }\sum_{i=1}^{\mathnormal{s}_{\mathnormal{j}}}|\omega_{i,k}^{j}| $ (so that layer $\mathnormal{L}_{\mathnormal{j}}$ satisfies $\left\| \mathbf{v}_{j}-\hat{\mathbf {v}}_{j} \right\| _\infty \leq \alpha \left\| \mathbf{v}_{j-1}-\hat{\mathbf {v}}_{j-1} \right\| _\infty$).

\vspace{1mm}
\noindent\textbf{CLEVER Score} Another robustness metric we adopt is the CLEVER score (Cross-Lipschitz Extreme Value for nEtwork Robustness)~\cite{Clever}, which is a recently proposed attack-independent robustness score for large scale networks. 

Given a sample input $x_{0}$ and a DNN $\nn$, we say $x_{a}$ is a perturbed example of $x_{0}$ with perturbation $\delta$ if $x_{a}=x_{0}+\delta$, let $\Delta_{p} = \|\delta\|_{p}$ denotes the $\ell_{p}$ norm of $\delta$, thus an adversarial example is a perturbed example $x_{a}$ that satisfy $\nn(x_{0})\not=\nn(x_{a})$, the minimum $\ell_{p}$ adversarial distortion of $x_{0}$, denoted as $\Delta_{p, \min }$, is defined as the smallest $\Delta_{p}$ over all adversarial examples of $x_{0}$. The idea is to approximately calculate the lower bound of $\Delta_{p, \min }$ of a given sample utilizing extreme value theory. The lower bound, denoted by $\beta_{L}$ where $\beta_{L} \leq \Delta_{p, \min }$, is defined such that any perturbed example of $x_{0}$ with $\|\delta\|_{p} \leq \beta_{L}$ are not adversarial examples. CLEVER score has been experimentally evaluated, which shows that it is consistent with other robustness evaluation metrics, e.g., attack-induced distortion metrics. Readers are referred to ~\cite{Clever} for details.


%% file: 4_Experiment.tex
\section{Methodology}
\label{sec:meth}



\begin{figure*}[t]
\begin{center}
\includegraphics[scale=0.55]{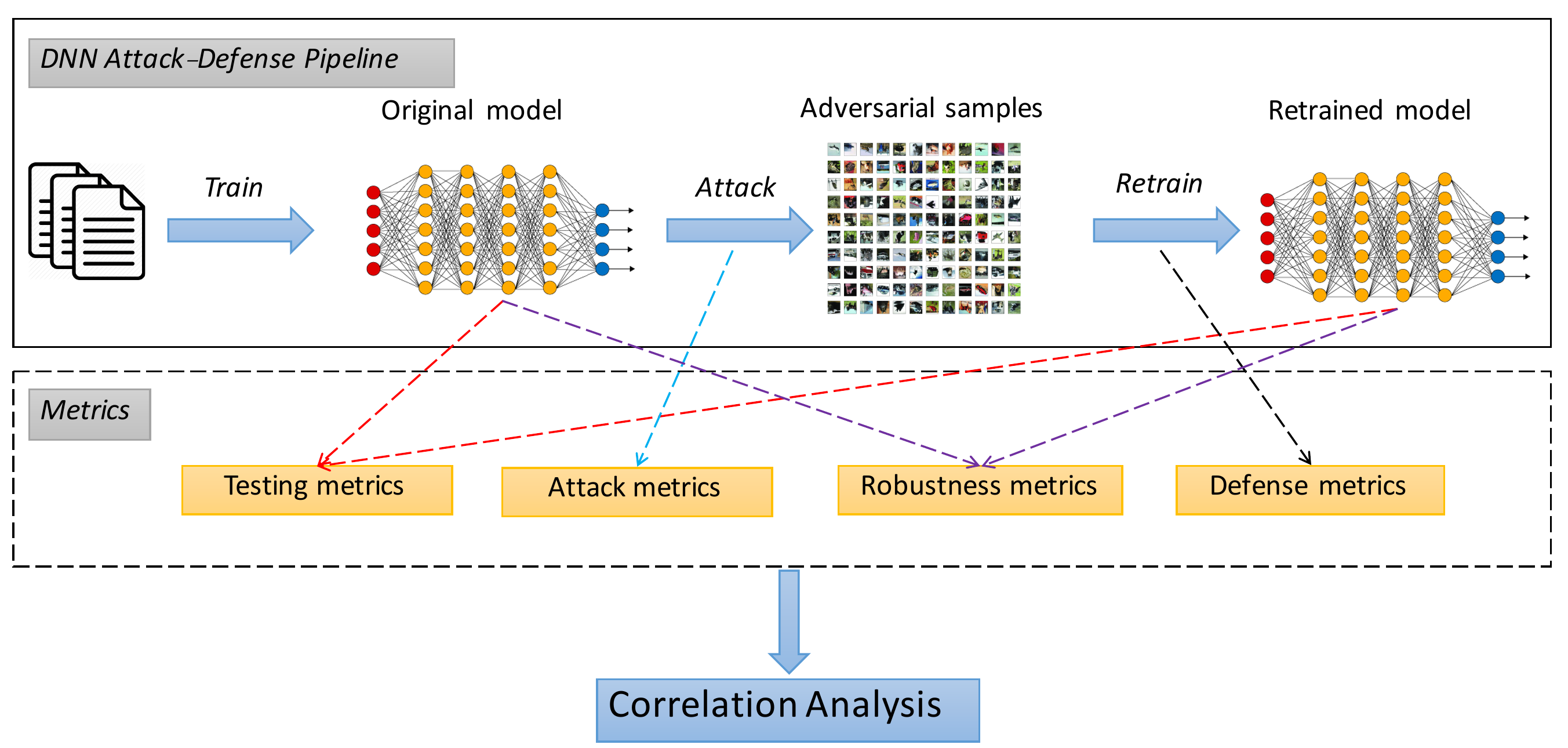}
\end{center}
\caption{Overview of experiment design}
\label{fig:overview}
\vspace{-4mm}
\end{figure*}

\subsection{Experiment Design}
The overall workflow of our experiment is shown in Figure~\ref{fig:overview}. We follow a common DNN testing process (e.g., by~\cite{deepxplore, DeepGauge}), as shown at the top of the figure, whilst extracting a variety of metrics (as shown in the middle of the figure) which are used for correlation analysis (as shown at the bottom). We start with training a model from a training set using state-of-the-art training methods. 
Afterwards, various adversarial attacks~\cite{fgsm, jsma, cw} are applied to generate new test cases. 
The last step is to augment the training set with the new test cases and obtain a retrained model. 

We collect four different groups of metrics to characterize different components of the process, i.e., (1) a set of testing coverage metrics of both the original and the retrained models, (2) a set of attack metrics of different kinds of adversarial attacks on the original models, (3) a set of robustness metrics of both the original models and the retrained models, and (4) a set of defense metrics which measure the differences between the retrained model and the original model. We repeat the above (attack and retrain) process for the $25$ seed models, obtain in total $100$ models, calculate the corresponding metrics and then conduct correlation analysis on all these metrics. In the following, we illustrate the challenges and our design choices of each part in detail.

\vspace{1mm}
\noindent \textbf{\emph{Adversarial Attacks}}
\label{sec:aegeneration}
We adopt three state-of-the-art DNN attack methods, i.e., FGSM~\cite{fgsm}, CW~\cite{cw} and JSMA~\cite{jsma}, which are introduced in section~\ref{sec:attackmethod}, to generate adversarial samples.
These attack methods are commonly used by previous DNN testing approaches, e.g.,~\cite{DeepGauge, SurpriseAdequacy}.
These generated adversarial samples are combined with the original datasets as new (training and testing) datasets for model retraining.
\vspace{1mm}

\noindent \textbf{\emph{Model Retraining}}
\label{sec:train}
For each original model, we obtain three sets of adversarial samples, one for each attack method. During model retraining, we combine the original training set with one set of the adversarial samples to obtain a new training set. We retrain the original model with the new training set to obtain a retrained model. As a result, we obtain $3$ retrained models for each original model, one for each attacking method.
We follow the standard partition of $6:1$ for training and testing on the MNIST dataset and $5:1$ for the CIFAR10 dataset.
\vspace{1mm}

\noindent \textbf{\emph{Metric Calculation}}
\label{sec:metric calculation}
As our objective is to investigate the correlations between coverage, robustness and other metrics associated with DNN, we conduct a thorough survey on existing metrics and collected \metricnum metrics in total. These metrics are categorized into four groups, i.e., testing metrics, robustness metrics, attack metrics and defense metrics. They are summarized in Table~\ref{tab:metric_name}. Note that the attack metrics measure to what extent the attacks are successful, imperceptible, whereas the defense metrics measure mainly on how the retrained models preserve the accuracy of the original model. For brevity, we refer the readers to the original papers for details. We calculate values of all metrics based on their original definitions and use default parameters according to their original papers.
\vspace{1mm}

\begin{table}[t]
\vspace{-1mm}
\caption{Summary of metrics}
\label{tab:metric_name}
\resizebox{.5\textwidth}{!}{
\begin{tabular}{|c|l|l|}
\hline
Metric Type & Metric Name & Description\\
\hline
\hline
\multirow{8}{*}{Testing} & NC   & Neuron Coverage ~\cite{deepxplore}\\
\cline{2-3} & KNC   & K-multisection Neuron Coverage ~\cite{DeepGauge}\\
\cline{2-3} & SNAC    & Strong Neuron Activation Coverage ~\cite{DeepGauge}\\
\cline{2-3} & NBC   & Neuron Boundary Coverage ~\cite{DeepGauge}\\
\cline{2-3} & TKNC    &Top-k Dominant Neuron Coverage ~\cite{DeepGauge}\\
\cline{2-3} & TKNP    & Top-k Dominant Neuron Patterns Coverage ~\cite{DeepGauge}\\
\cline{2-3} & LSA/DSA    & Surprise adequacy to training set ~\cite{SurpriseAdequacy}\\
\hline
\multirow{2}{*}{Robustness} & Lipschitz constant  & The global Lipschitz constant ~\cite{Robustness}\\
\cline{2-3} & CL1/CL2/CLi  & Clever score with L$_1$/L$_2$/L$_\infty$ norm ~\cite{Clever}\\
\hline
\multirow{7}{*}{Attack} & MR  & Misclassification Ratio ~\cite{DeepSec}\\
\cline{2-3} & ACAC  & Average Confidence of Adversarial Class ~\cite{DeepSec}\\
\cline{2-3} & ACTC & Average Confidence of True Class ~\cite{DeepSec}\\
\cline{2-3} & ALD$_p$ & Average $L_p$ Distortion ~\cite{DeepSec}\\
\cline{2-3} & ASS & Average Structural Similarity ~\cite{ass}\\
\cline{2-3} & PSD & Perturbation Sensitivity Distance ~\cite{psd}\\
\cline{2-3} & NTE & Noise Tolerance Estimation ~\cite{psd}\\
\cline{2-3} & RGB & Robustness to Gaussian Blur ~\cite{DeepSec}\\
\cline{2-3} & RIC & Robustness to Image Compressionr ~\cite{DeepSec}\\
\cline{2-3} & CC & Computation Cost ~\cite{DeepSec}\\
\hline
\multirow{4}{*}{Defense} & CAV  & Classification Accuracy Variance ~\cite{DeepSec}\\
\cline{2-3} & CRR/CSR  & Classification Rectify/Sacrifice Ratio ~\cite{DeepSec}\\
\cline{2-3} & CCV & Classification Confidence Variance ~\cite{DeepSec}\\
\cline{2-3} & COS & Classification Output Stability ~\cite{DeepSec}\\
\hline

\end{tabular}
}
\vspace{-6mm}
\end{table}

\noindent \textbf{\emph{Correlation Analysis}}
\label{sec:correlation}
We conduct correlation analysis, a statistical technique that shows whether and how strongly pairs of variables are correlated, on the metrics. We are particularly interested to observe which metrics are correlated to the robustness of a DNN model. 
The resulting correlation coefficient is a single value between $-1$ and $+1$, where $+1$ (and $-1$) means the most positively (and negatively) correlated, and $0$ means no correlation. In this work, we adopt a commonly used correlation coefficients, Kendall's $\tau$ rank correlation coefficient~\cite{Kendall}, which is a rank based correlation that measures monotonic relationship between two variables, to measure the correlations between different metrics. Note that compared to alternative methods like Pearson product-moment correlation coefficient~\cite{Pearson}, Kendall's $\tau$ rank correlation coefficient does not require that the dataset follows a normal distribution or the correlation is linear.
%
Since we adopt two popular dataset MNIST and CIFAR10 to train $4$ different families of DNN models. We calculate the correlations of different metrics for the two dataset separately, in order to avoid the potential impact due to the training data.

%% file: 4_Implementation.tex
\begin{figure}[t]
\begin{center}
\includegraphics[scale=0.28]{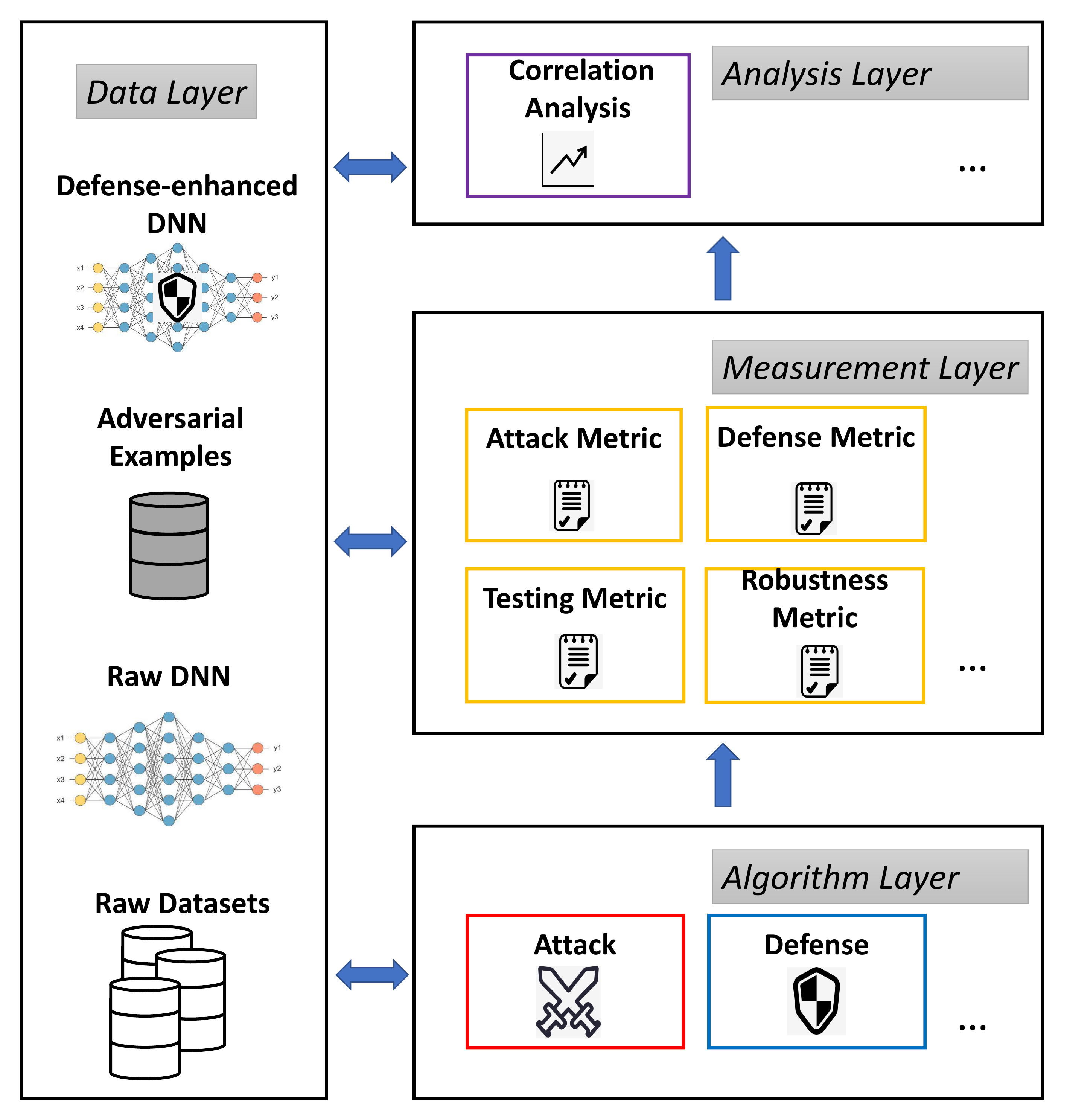}
\end{center}
\label{fig:architecture}
\caption{The architecture of our framework}
\label{fig:arch}
\vspace{-8mm}
\end{figure}

\section{Implementation and Configurations}
\label{sec:implementation}

Our system is implemented based on the TensorFlow framework~\cite{tensorflow} and the architecture is shown in Figure~\ref{fig:arch}.
There are $4$ layers, i.e., the data layer, the algorithm layer, the measurement layer and the analysis layer. Our implementation is designed to be extensible, i.e., each layer can be extended with new models and algorithms with little impact on the other layers. 
Our implementation, including all the data and algorithms, is open source on GitHub\footnote{https://github.com/icse-2020/DRTest}.

\vspace{1mm}
\noindent \textbf{\emph{The data layer}} maintains all data used in our study, including the original models, the adversarial samples generated for original models, and the retrained models. It interacts with all other layers. We use well-know models on image classification tasks in our experiment. To cover a range of different deep learning model structures, we adopt four different model families, including $3$ LeNet family models (LeNet-1, LeNet-4 and LeNet-5~\cite{lenet}), $4$  VGG family models (VGG-11, VGG-13, VGG-16, VGG-19~\cite{vgg}), $4$  ResNet family models (ResNet-18, ResNet-34, ResNet-50, ResNet-101~\cite{ResNet}) and $3$ GoogLeNet family models (GoogLeNet-12, GoogLeNet-16, GoogLeNet-22~\cite{googlenet}). In total, we have $14$ model structures, which are representative image classification models.

\begin{table}[t]
\scriptsize
\caption{Hyper-parameters of attack methods}
\label{tab:attack}
\begin{tabular}{|c|c|c|c|c|}
\hline
dataset & attack method & model family & parameter & success rate\\
\hline
\hline
\multirow{12}{*}{MNIST} & \multirow{4}{*}{FGSM} & LeNet & \multirow{4}{*}{0.2, 0.3, 0.4} & 0.94\\
\cline{3-3}
\cline{5-5}
& & VGG & & 0.81\\
\cline{3-3}
\cline{5-5}
& & ResNet & & 0.83\\
\cline{3-3}
\cline{5-5}
& & GoogLeNet & & 0.71\\
\cline{2-5}

& \multirow{4}{*}{CW} & LeNet & \multirow{4}{*}{9, 10, 11} & 0.91\\
\cline{3-3}
\cline{5-5}
& & VGG & & 0.81\\
\cline{3-3}
\cline{5-5}
& & ResNet & & 0.91\\
\cline{3-3}
\cline{5-5}
& & GoogLeNet & & 0.90\\
\cline{2-5}
& \multirow{4}{*}{JSMA} & LeNet & \multirow{4}{*}{0.09, 0.1, 0.11} & 0.89\\
\cline{3-3}
\cline{5-5}
& & VGG & & 0.25\\
\cline{3-3}
\cline{5-5}
& & ResNet & & 0.75\\
\cline{3-3}
\cline{5-5}
& & GoogLeNet & & 0.52\\
\hline
\multirow{9}{*}{CIFAR} & \multirow{3}{*}{FGSM} & VGG & \multirow{3}{*}{0.01, 0.02, 0.03} & 0.76\\
\cline{3-3}
\cline{5-5}
& & ResNet & & 0.65\\
\cline{3-3}
\cline{5-5}
& & GoogLeNet & & 0.75\\
\cline{2-5}

& \multirow{3}{*}{CW} & VGG & \multirow{3}{*}{0.1, 0.2, 0.3} & 0.88\\
\cline{3-3}
\cline{5-5}
& & ResNet & & 0.90\\
\cline{3-3}
\cline{5-5}
& & GoogLeNet & & 0.90\\
\cline{2-5}
& \multirow{3}{*}{JSMA} & VGG & \multirow{3}{*}{0.09, 0.1, 0.11} & 0.80\\
\cline{3-3}
\cline{5-5}
& & ResNet & & 0.79\\
\cline{3-3}
\cline{5-5}
& & GoogLeNet & & 0.75\\
\hline
\end{tabular}
\vspace{-4mm}
\end{table}

We adopt two popular publicly-available datasets, i.e., MNIST~\cite{MNIST_Data} and CIFAR10~\cite{Cifar} to train DNN models in our work. MNIST is a set of handwritten digit images. It contains $70,000$ images in total. Each image in MINIST dataset is single-channel of size $28 * 28 * 1$. CIFAR10 is a set of color images. It contains $10$ classes, each of which has $6,000$ images, and the input size of each image is $32 * 32 * 3$.

\vspace{1mm}
\noindent \textbf{\emph{The algorithm layer}} contains a set of algorithms for attacking DNN as well as algorithms for defending DNN through retraining. For each trained model, we adopt three state-of-the-art attack methods (e.g., FGSM, CW and JSMA) to generate adversarial samples. The principle of choosing parameters for each attack is to balance the imperceptibility and success rate of generating adversarial samples. For MNIST, we adopt the same parameters from cleverhans~\cite{cleverhans} for all three attacks. For CIFAR10, we slightly changed the parameters of FGSM and CW in order to obtain better imperceptibility. The parameters chosen include the attack step size for FGSM, the initial tradeoff-constant for tuning the relative importance of size of the perturbation and confidence of classification for CW and the maximum percentage of perturbed features for JSMA.

To further avoid bias introduced by hyper-parameters, we run each attack method on the original dataset for $3$ times, each time with a different hyper-parameter configuration. Then we combine the successful adversarial samples generated from $3$ runs of attacks as the adversarial sample set for model retraining. Table~\ref{tab:attack} shows the details of the hyper-parameter configurations for each attack method, and the column \texttt{hyperparameter} summarizes the hyper-parameter configurations used in each run of attack.


During training and retraining, we adopt a learning rate of $0.001$, a batch size of $128$ for all models in the two datasets. For MNIST, a test accuracy above $98\%$ is accepted in both training and retraining. For CIFAR10, a test accuracy above $80\%$ is accepted during training process and a test accuracy above $85\%$ is required for retraining.

\vspace{1mm}
\noindent \textbf{\emph{The measurement layer}} contains all implementation for calculating the \metricnum metrics shown in Table~\ref{tab:metric_name}. We calculate four robustness values, i.e., Global Lipschitz Constant (Lipz) and the CLEVER score (CL1, CL and CLi) for each model. Note that LeNet is not feasible for CIFAR10. In our experiment, since calculate CLEVER score is extremely time-consuming for GoogLeNet, we reduce the number of images to $50$ and sampling parameter $N_{b}=50$, as it is reported that $50$ or $100$ samples are usually sufficient to obtain a reasonably accurate robustness estimation~\cite{Clever}.
We calculate the coverage criteria of different DNN models with the same test suite (i.e., the original test suite of MINIST or CIFAR10) and obtain \nummodel$*4$ and $11*4$ values of each coverage criteria on MNIST and CIFAR10, respectively.

Defense Metrics are calculated for all the defense enhanced models, i.e., models after adversarial training, according to their original definitions~\cite{DeepSec}. For each dataset, We obtain \nummodel$*3$ and $11*3$ values for each defense metric on MINIST and CIFAR10, respectively.
Attack Metrics are calculated for the generated adversarial examples of each attack method, all parameters of attack metrics are set based on  their original definitions~\cite{DeepSec}. We obtain \nummodel$*3$ and $11*3$ values for each attack metric on MINIST and CIFAR10, respectively.

We additionally calculate a set of $\Delta$-metrics, which are denoted as Metric-diff. For instance, Lipz-diff is the Lipschitz Constant of the retrained model minus that of the original model. We obtain \nummodel$*3$ and $11*3$ $\Delta$-robustness for each robustness metric on MINIST and CIFAR10. Similarly, we calculate $\Delta$-coverage metrics by subtracting the coverage achieved by the augmented test set (i.e., the original test set plus the adversarial samples) from that of the original test set. We obtain \nummodel$*3$ and $11*3$ $\Delta$-coverage values for each coverage metric on MINIST and CIFAR10.

%
%
%
%

\vspace{1mm}
\noindent \textbf{\emph{The analysis layer}} implements the correlation analysis algorithm~\cite{Kendall}. We first plot the data to observe the trend and then decide on the correlation analysis method to use. By observing the data plot, we found that the data does not show a linear trend. Therefore, we choose the Kendall's $\tau$ rank correlation coefficient~\cite{Kendall}, which does not assume that the data follows a normal distribution or the variables have a linear correlation.

\vspace{1mm}
\noindent All experiments were conducted using four GPU servers. Server 1 has 1 Intel Xeon 3.50GHz CPU, 64GB system memory and 2 NVIDIA GTX 1080Ti GPU. Server 2 has 2 Intel Xeon 2.50GHz CPU, 126GB system memory and 4 NVIDIA GTX 1080Ti GPU. Server 3 has 2 Intel Xeon 2.50GHz CPU, 96GB system memory and 4 NVIDIA GTX 1080Ti GPU. Server 4 has 1 Intel Xeon 2.50GHz CPU, 119GB system memory and 2 Tesla P100 GPU. Not all GPUs on the four servers are fully utilized. We remark that we do not always have full occupations  of all GPUs and 6 GPUs are used on average during the experiment period. 

In total, the experiment took more than $6,100$ GPU hours to finish. Table~\ref{tab:time} shows the time spent on different steps, i.e., on generating adversarial examples, training and retraining, as well as metric calculations for each dataset on each model. The unit is $1$ GPU hour. The time for correlation calculation compared to the other steps is neglectable. The most time consuming step is the metric calculation, which took $1,350$ hours for the ResNet family on CIFAR10. The most time consuming metrics is the coverage criteria, which varies significantly depending on the model structure. Adversarial sample generation is also time consuming.

\begin{table}[t]
\scriptsize
\caption{Time for different steps in the experiment}
\label{tab:time}
\begin{tabular}{|c|c|c|c|c|}
\hline
dataset & model family & generate AE & train \& retrain & metric calc\\
\hline
\hline
\multirow{4}{*}{MNIST} & LeNet & \textless 0.5  & \textless 0.5  & \textless 0.5 \\
\cline{2-5}
& VGG & 160 & 6  & 420 \\
\cline{2-5}
& ResNet & 240  & 12  & 1200 \\
\cline{2-5}
& GoogLeNet & 120  & 25  & 300 \\
\hline
\multirow{3}{*}{CIFAR10} & VGG & 540  & 12  & 550 \\
\cline{2-5}
& ResNet & 450  & 45  & 1350  \\
\cline{2-5}
& GoogLeNet & 300  & 50  & 330 \\
\hline
\end{tabular}
\end{table}

%% file: 5_Discussion.tex
\section{Findings}
\label{sec:exp}

\input{findings}

%% file: findings.tex
\subsection{Research Questions}


\vspace{1mm}
\noindent\textbf{RQ1: Are there any correlations between existing test coverage criteria and the robustness of the DNN models?}
%


\begin{figure*}[t]
\centering
\small
\begin{subfigure}{0.45\textwidth}
     \includegraphics[width=\linewidth]{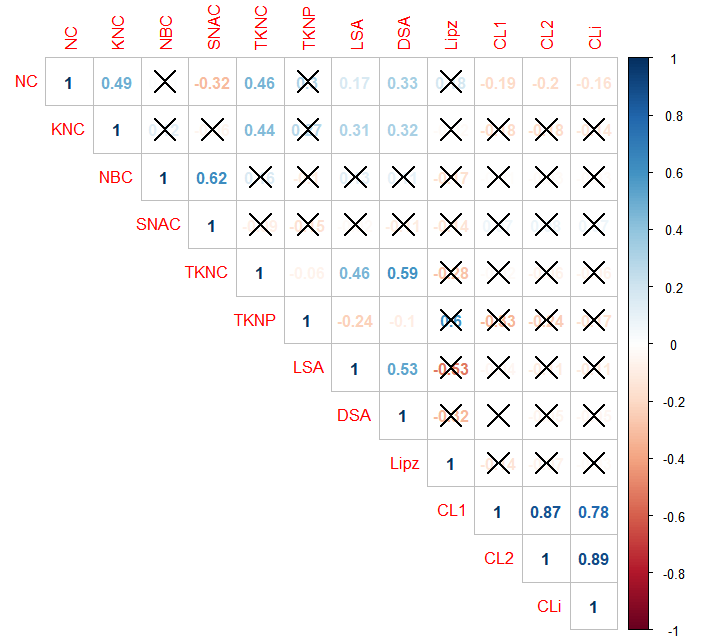}
     \caption{MNIST}
     \label{fig:1_mnist_kendall_all}
\end{subfigure}
\begin{subfigure}{0.45\textwidth}
     \includegraphics[width=\linewidth]{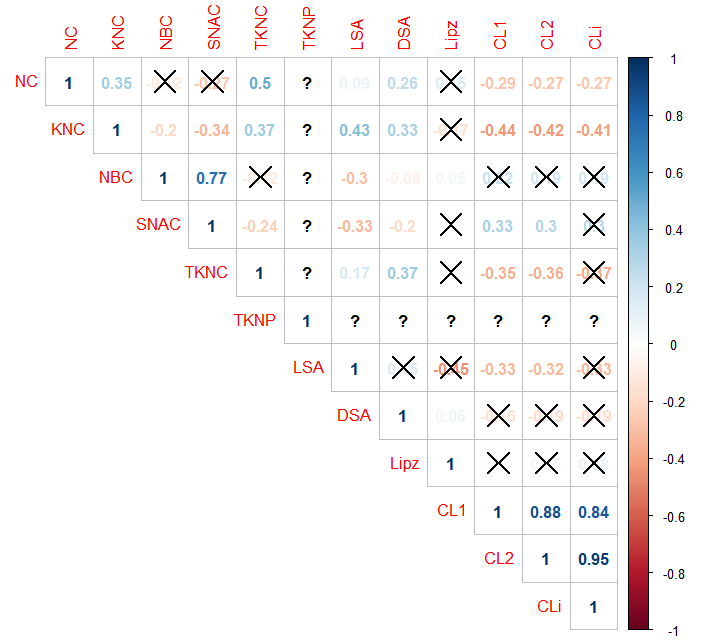}
     \caption{CIFAR10}
     \label{fig:1_cifar_kendall_all}
\end{subfigure}
\caption{Test coverage vs. robustness metrics }
\label{fig:kendall_r_cc}
\vspace{-4mm}
\end{figure*}


To answer the question, we conduct correlation analysis on the coverage metrics and the robustness metrics of all models on the original test set. The results are shown in Figure~\ref{fig:kendall_r_cc}.
The number and the color represent the strength of the correlation.
The correlation value is a number between $-1$ and $1$. Positive number (and blue color)  indicates positively correlated and negative number (and red color) indicates negative correlated. The larger the absolute number is, the stronger the correlation is. The darker the color is, the stronger the correlation is.
We measure the p-value of the sample data set we have and regard p-value greater than $0.05$ as insignificant. An ``X'' mark means that we cannot make a decision because p-value is larger than $0.05$ (i.e., insignificant) and a question mark ``?'' means that there are no valid results since the standard variation of the data is $0$. The same notations are used in subsequent figures as well. We summarize the results in the following two aspects.
According to the definition of correlation
in Guildford scale~\cite{Guildford-scale}, an absolute value of less than $0.4$ means that the (positive or negative) correlation is low; an absolute value of $0.4$ - $0.7$ means that the correlation is
moderate; and otherwise the correlation is high or very high (i.e., $0.7$-$0.9$ or above $0.9$, respectively).

We have the following observations based on Fig.~\ref{fig:kendall_r_cc}. First, there is no significant or negative correlation between coverage and robustness metrics. In particularly, neural coverage is negatively correlated (i.e., with a value between $-0.16$ and $-0.29$) with the CLEVER score and is not significantly correlated with Lipschitz constant for both MNIST and CIFAR10. Moreover, KNC, TKNC and LSA also show negative correlations with CLEVER score on CIFAR10. It suggests that a DNN is less robust if the test set has a larger neuron coverage (although the strength of the correlation is weak), which is unexpected.
Second, there is no significant correlation between any of the other coverage and any of the robustness metrics on the MNIST dataset. For the CIFAR10 dataset, positive correlation is only observed between SNAC and the CLEVER score, and the strength is low.  
This result suggests that a DNN model which achieves high coverage is not necessary robust and vice versa.


We further investigate the correlation among all test coverage criteria themselves. It can be observed from Fig.~\ref{fig:kendall_r_cc} that NC, KNC, TKNC, LSA and DSA are positively correlated with each other. NBC and SNAC are correlated with each other with medium or high strength, whereas they have no (or weak negative) correlation with the other metrics. The results are consistent with observations reported in~\cite{DeepGauge} and~\cite{SurpriseAdequacy} which propose these coverage. This suggests that despite that different coverage criteria are defined differently, they are in general correlated (except for the boundary coverage).

We have the following answer to RQ1.

\begin{framed}
\vspace{-2mm}
\noindent \emph{Different coverage criteria are correlated with each other. There is limited correlation between the coverage criteria and the robustness metrics.}
\vspace{-2mm}
\end{framed}

\begin{figure*}[t]
\centering
\small
\begin{subfigure}{0.47\textwidth}
    \includegraphics[width=\linewidth]{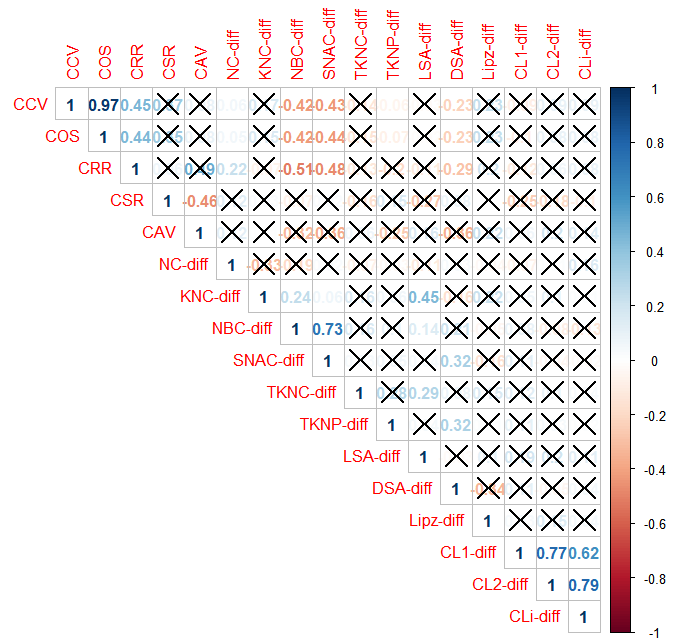}
     \caption{MNIST}
     \label{fig:2_mnist_kendall_all}
\end{subfigure}
\begin{subfigure}{0.47\textwidth}
    \includegraphics[width=\linewidth]{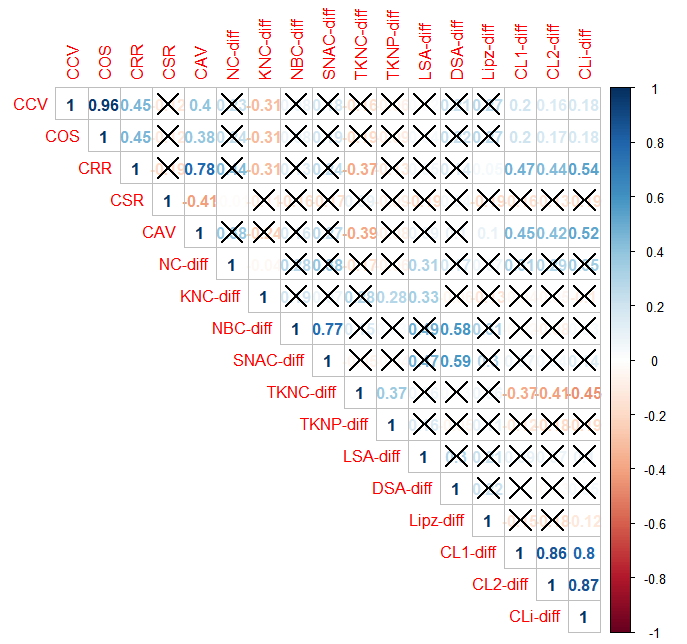}
     \caption{CIFAR10}
     \label{fig:2_cifar_kendall_all}
\end{subfigure}
\caption{Defense Metrics vs. Coverage Criteria  Differences vs.  Robustness Differences}
\label{fig:kendall_cdiff_rdiff_def}
\vspace{-4mm}
\end{figure*}

\vspace{1mm}
\noindent\textbf{RQ2: Does retraining with new test cases which improves coverage criteria improve the robustness of a DNN model?}

To answer this question, we conduct correlation analysis on the difference on coverage criteria and the difference on robustness metrics before and after retraining. The results are shown in Fig.~\ref{fig:kendall_cdiff_rdiff_def}. We observe that there is no correlation between the difference on any coverage criteria and the difference on any  robustness metrics, except that there is \emph{negative} correlation between TKNC-diff and the CLEVER scores for all the CIFAR10 models. \emph{This result casts a shadow over existing testing approaches, as the existing testing approaches are designed to generate test cases for high coverage, with the hope that such test cases can be used to improve the adversarial robustness of the DNN models.}

We thus have the following answer to RQ2.
\begin{framed}
\vspace{-2mm}
\noindent \emph{Retraining with new test cases which improve the coverage criteria does not necessarily improve the model robustness.
}
\vspace{-3mm}
\end{framed}

\begin{figure*}[t]
\centering
\small
\begin{subfigure}{0.47\textwidth}
    \includegraphics[width=\linewidth]{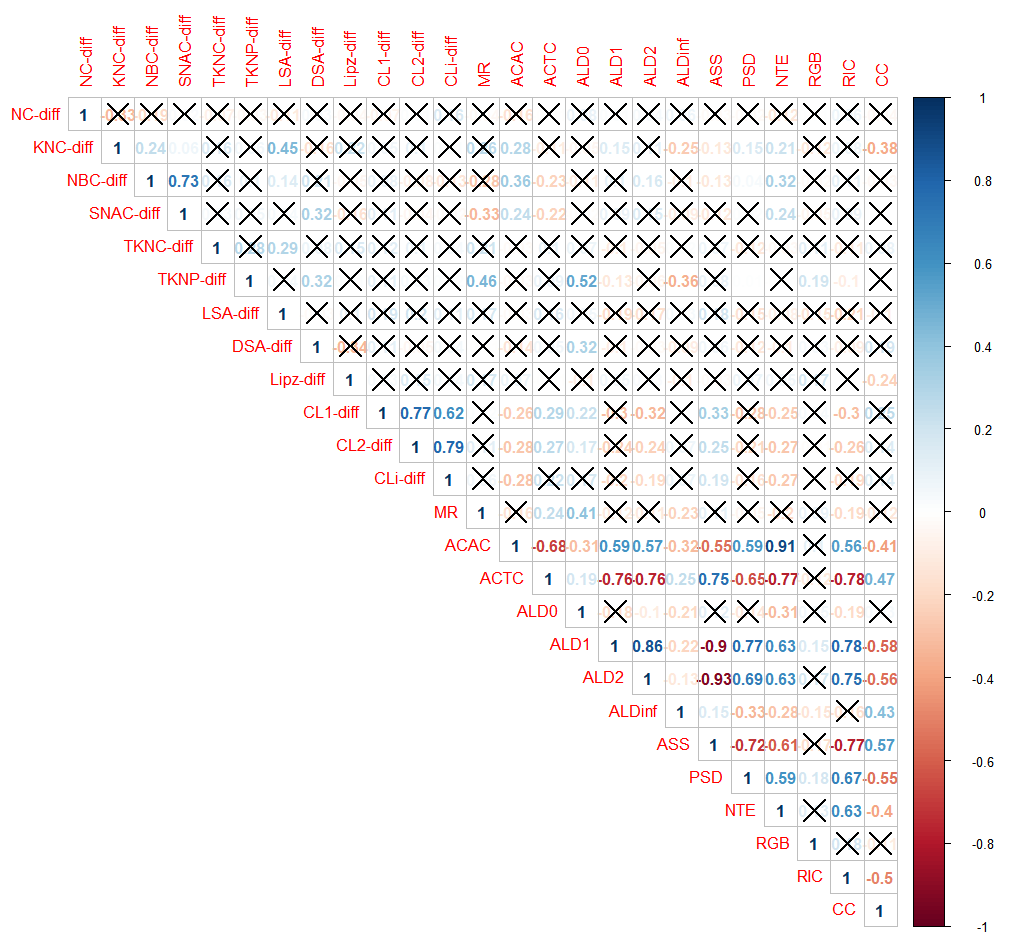}
     \caption{MNIST}
     \label{fig:3_mnist_kendall_all}
\end{subfigure}
\begin{subfigure}{0.47\textwidth}
    \includegraphics[width=\linewidth]{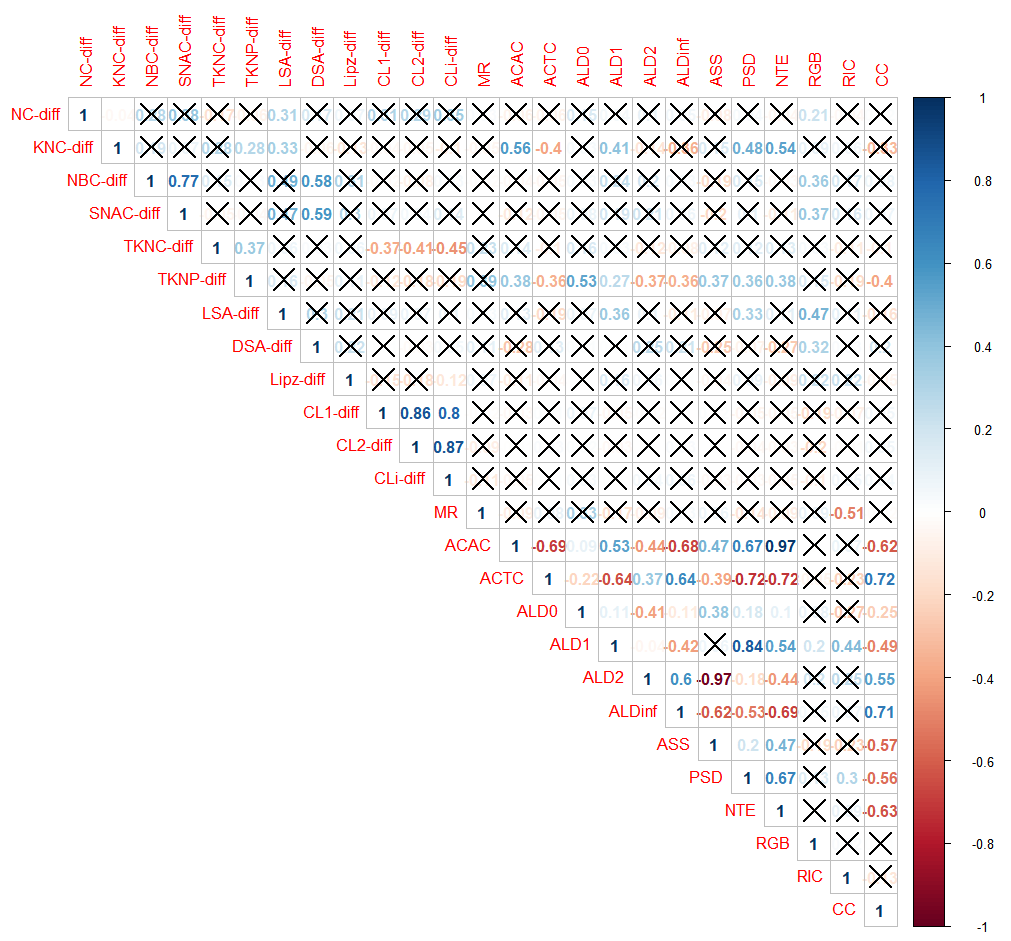}
     \caption{CIFAR10}
     \label{fig:3_cifar_kendall_all}
\end{subfigure}
\caption{Coverage Criterion Difference vs. Robustness Difference  vs. Attack Metric}
\label{fig:kendall_ccdiff_rdiff_att}
\vspace{-4mm}
\end{figure*}

\vspace{1mm}
\noindent\textbf{RQ3: Are there metrics that are strongly correlated to the improvement of model robustness?}

The above results show that existing test coverage criteria have limited correlations with the robustness of DNN models and testing methods based on improving the coverage do not improve the robustness of DNN models. The question is then: are there metrics which are correlated to the improvement of the model robustness? To answer the question, we systematically conduct correlation analysis between all metrics (or the metrics's difference before and after retraining) and the improvement of the model robustness. 




The correlations between the defense metrics and the improvement of robustness are shown in Fig.~\ref{fig:kendall_cdiff_rdiff_def}. We observe that there are positive correlations between the difference of the CLEVER scores and all the defense metrics on CIFAR10. In particular, the correlation is of medium level for CRR and CAV. CRR and CAV measure how much the defense-enhanced model preserves the functionality of the original model~\cite{DeepSec}. Intuitively, this indicates that \emph{a defense method leads to more robustness improvement if the original model is better preserved by the defense-enhanced model.}
Furthermore, given the huge cost on computing robustness metrics, such positive correlations potentially provide a lightweight way of estimating on the effectiveness of a model enhancement method.

We additionally analyze the correlation between the attack metrics and the improvement of coverage criteria. We have the following observations from the results shown in Fig.~\ref{fig:kendall_ccdiff_rdiff_att}. There are correlations between the differences of TKNP and KNC and the attack metrics.
Furthermore, NTE is positively correlated with KNC-diff, NBC-diff and SNAC-diff. RGB is positively correlated with NC-diff, NBC-diff and SNAC-diff. Intuitively, NTE and RGB measure the robustness of adversarial samples, which implies that more robust adversarial samples contribute more to the improvement of coverage metrics. Lastly, there is no correlation between the robustness differences and the attack metrics for the CIFAR10 dataset. For the MINIST dataset, we observe negative correlations between the CLEVER score differences with ACAC, ALD2, RIC and NTE, and positive correlations with ASS and ACTC. These observations indicate that \emph{more confident, perceptible and robust adversarial samples contribute more to improving the coverage criteria.}

We have the answer to RQ3.
\begin{framed}
\vspace{-2mm}
\noindent \emph{Some defense metrics are positively correlated to the improvement of model robustness.}
\vspace{-2mm}
\end{framed}

\vspace{1mm}
\noindent\textbf{RQ4: Are the correlation results consistent across different datasets, model families and correlation analysis methods?}

This question examines whether the correlation results are universal or rather may vary cross different datasets, model families or correlation analysis methods. To answer this question, we systematically conduct the different correlation analysis using data obtained from different datasets and model families. For the sake of space, we omit the details and refer the readers to the supplementary materials made available at the online repository for details.

Overall, while the correlation between testing coverage and robustness on MNIST and CIFAR10 are mostly consistent, we do observe that the results on some correlations vary slightly across the two datasets. For instance, the attack metrics (except ALDinf) show correlation with CL-diff on MNIST but not on CIFAR10. The defense metrics show strong correlation with robustness and robustness-diff on CIFRA10, which is not the case on MNIST.

There are also inconsistent correlation results across different model families. The correlation results on the MNIST, LeNet and VGG families are consistent, which is expected since they have similar model structures. However, it is surprise that models in the GoogLeNet family often show opposite correlation results to those of the MNIST, LeNet and VGG families, especially for correlation between the attack metrics and the improvement of the model robustness. This can be explained as GoogLeNet has a rather different architecture from MNIST, LeNet and VGG (GoogLeNet tends to have more neurons in a layer instead of having more layers).

The above-mentioned inconsistency suggests that the correlation may depend on the dataset and, more noticeably, the model architecture, which further complicates the picture.

Lastly, we apply different correlation analysis algorithms (including Pearson product moment correlation~\cite{Pearson} and Spearman's rank-order correlation~\cite{spearman}) to observe whether the results are consistent. Overall, although the results are not identical, the differences are not significant and the results (e.g., whether it is positively or negatively correlated or whether it is strongly or weakly correlated) remain consistent. We choose to present the results of Kendall correlation coefficients in this work as it requires the least assumption on the underlying data. The results of other correlation analysis algorithms are present in the supplementary materials online.

We have our answer to RQ4.
\begin{framed}
\vspace{-2mm}
\noindent \emph{The correlation results are consistent across different correlation analysis algorithms but may vary across different datasets or model families.}
\vspace{-2mm}
\end{framed}

\subsection{Explanation}

In the following, we aim to interpret and `explain' the above-mentioned results. These explanations must, however, be taken a grain of salt as they should be properly examined in the future. 

First, the reason that existing coverage criteria are not correlated with robustness may simply be due to the fact these coverage criteria are too weak to differentiate robust and not-robust DNN models. It has been shown that high neuron coverage could be easily achieved with a small number of samples~\cite{DeepCover}, and similar conclusions are given by Odena et al.~\cite{tensorfuzz} for coverages proposed in DeepGauge, such as neuron boundary coverage. This finding is confirmed by another recent research work~\cite{ccmisleadingICSE19}, which reports that adversarial examples are pervasively distributed in the space divided by coverage criteria. The work~\cite{ccmisleadingICSE19} also suggests that using structural coverage to measure the neural network robustness can be questionable.  

Second, our results suggest that retraining with the test case does not necessarily improve robustness. For software systems, a test case which reveals a bug naturally leads to bug fixing, which ``definitely'' improves the `robustness' of the system. This is not certain for DNN models. because the retrained model could be rather different from the original model, i.e., it is like a new model, due to how such models are trained (i.e., through optimization techniques which embody a lot of non-determinism and carry little theoretical guarantee).

Third, we consider it to be intuitive that defense metrics are correlated with robustness as these defense metrics are indeed less formal ways of measuring robustness (i.e., in term of how well a DNN model defends adversarial attacks). 

As for the answer to RQ4, we take the consistency between different correlation analysis algorithms positively as it shows that our results are not the result of certain `biased' correlation analysis algorithm. The second part of the answer may suggest that a testing method may have to be tailored according to different DNN architectures.   

\subsection{Discussion} 
The results discussed so far are mostly negative, i.e., only several defense metrics are correlated with the improvement of model robustness and existing testing methods designed based on coverage have limited effectiveness on improving the robustness of the DNN models. The results question the usefulness of coverage criteria proposed for DNN models. Indeed, a well tested (and improved by retraining) DNN through existing testing methods might produce a new model which has higher empirical accuracy on the testing set. However, the new model is not necessarily more robust than the original model against adversarial perturbations. In fact, a recent finding shows that \emph{DNN model robustness maybe at odds with accuracy since robust classifiers are learning fundamentally different feature representations than standard classifiers~\cite{tsipras2018robustness}}. 
For DNN models to be deployed in safety-critical applications, we believe that robustness is an as (if not more) important property as accuracy. The real question thus remains: \emph{how should we test DNN models and make use of the testing results so that the robustness of the DNN models is improved? Or are there ways to improve the robustness of the DNN models in general?}.

To this question, we do not have a clear answer and thus it remains an open question to us. It is possible that there could be other coverage criteria which are correlated with the model robustness or the associated testing method can help improve the model robustness. It is however important that no matter what coverage is proposed, it must be thoroughly analyzed to show its effect on model robustness.

Our view is that finding adversarial samples should not be the end of DNN testing. Rather, testing DNN models should be designed in consideration of the model enhancement methods, i.e.,  a testing method should produce test cases which are useful according to the model enhancement methods. For instance, given the positive correlation between robustness and the defense metrics, we might want to generate test cases which could contribute to improve defense metrics such as CAV and CCV.

\subsection{Threats to validity} 

First, there may be threats to validity due to the selected datasets and model structures. In this work, we regard each DNN model as the a program of the same functionality and calculate different metrics on these models. We assume the metrics are valid across different DNN model structures and conduct correlation analysis on the obtained metrics. However, some metrics are not applicable to certain model structures (e.g., MC/DC is not applicable to ResNet and GoogLeNet).
Besides, Since each model family has limited number of models and datasets to analyse with, the results may be biased to these specific datasets and model structures even though we are adopting the most popular datasets and state-of-the-art model structures.

Second, there may be threats to validity due to the limited size of datasets, models and attack methods adopted. In this work, we use $14$ different DNN model structures, $3$ adversarial attack methods, $100$ models, $2$ datasets, and \metricnum different metrics. While we are working on more datasets, model structures, etc., we could not significantly increase the scale due to the huge cost (more than $6,100$ GPU hours) of the empirical study.   For more statistical significant results, more data points are helpful (or even necessary). We thus call upon the open source community to jointly upscale our study. To make sure that our correlation analysis results are valid, we only report the results beyond a certain significant level by measuring its p-value~\cite{p-value} in this work.

Third, the evaluation of DNN model robustness in general is still an open and challenging research problem~\cite{zhang2018adversarial}. Although we are adopting the most popular robustness metrics, there might still be threat to validity to what extent these metrics can actually reflect the robustness of the models.

%% file: 2_Related_Work.tex
\section{Related Work}
\label{sec:related}
In this section, we review related works, with a focus on recent progress on 1) testing approaches which propose different testing criteria for DNN models, 2) different robustness metrics to evaluate the quality of the DNN models, and 3) state-of-the-art adversarial attacks and defense methods.

\vspace{1mm}
\noindent\textbf{Testing of deep learning models}
Several recent papers proposed different coverage criteria for evaluating the effectiveness of a test set, along with different methods to generate test cases to improve the coverage criteria. For instance, DeepXplore~\cite{deepxplore} proposed the first testing criterion for DNN models, i.e., Neuron Coverage (NC), which calculates the percentage of activated neurons (w.r.t. an activation function) among all neurons. Later, DeepGauge~\cite{DeepGauge} extended the idea and proposed a serial of more fine-grained multi-granularity testing criteria from both neuron level and layer level. Inspired by the MC/DC test criteria from traditional software testing, Sun et al. proposed four test criteria based on syntactic connections between neurons in adjacent layers and a concolic testing strategy to systematically improve MC/DC coverage of DNN models~\cite{DeepConcolic}. More recently, two surprise adequacy criteria~\cite{SurpriseAdequacy} are proposed to measure the level of `surprise' of a new test case to the training set, e.g., by measuring the distance between their activation vectors. Our work implemented and reviewed most of the above-mentioned coverage criteria for a comprehensive evaluation. Note that some are omitted as they are extremely costly to compute.

\vspace{1mm}
\noindent\textbf{Robustness of deep learning models}
In the machine learning and the formal verification community, multiple metrics are used to measure the robustness of DNN models. The Lipschitz constant was proved to be useful as a metric for Feed-forward Neural Networks by Xu, H.~\cite{Robustness}. Segedy et al.~\cite{Intriguing} leveraged the product of Lipschitz constants for each layer as a measure of the DNN robustness and proposed Parseval Networks~\cite{Parseval} to achieve improved robustness by maintaining a small Lipschitz constant at every hidden layer. Adversarial manipulation, which looks at the required distortion of adversarial samples is another direction. Matthias et al. intended to gave a formal guarantee on the robustness of a classifier by obtaining a robustness lower bound using a local Lipschitz continuous condition~\cite{Formalguarantees}. Recently, Weng et al.~\cite{Clever} extended their work and proposed a robustness metric called CLEVER score which is calculated using extreme value theory. Our work adopted one latest criteria from each direction.

\vspace{1mm}
\noindent\textbf{Attack and Defense for deep learning models}
There is a large body work on adversarial attack and defense in recent years, which we are only able to cover the most relevant ones. In particular, we adopted three state-of-the-art attacks to generate adversarial samples, i.e., a gradient-based approach (the FGSM method~\cite{fgsm}), a saliency map-based approach (JSMA~\cite{jsma}), and an optimization-based approach (C\&W attack~\cite{cw}). On the defense side, multiple attempts are available to obtain a relatively robust model at training phase or detect adversarial samples at runtime. For instance, adversarial training tries to include adversarial samples into consideration~\cite{scale}. Another relevant direction is robust training which tries to train a robust DNN model by considering all the possible perturbation at training phase~\cite{madry}. Besides, mutation testing is adopted to find adversarial samples at runtime~\cite{ouricse19}. Essentially, testing is complementary to these defense works.


%% file: 6_Conclusion.tex
\section{Conclusion}
\label{sec:con}
In this work, we conducted a systematic and quantitative empirical study on $100$ state-of-the-art DNN models to investigate the relevance and effectiveness of recently proposed testing criteria and approaches for deep neural networks. Our study is based on a self-contained toolkit which implements all the testing coverage criteria, two robustness metrics and a large set of measurable metrics during the adversarial attack and defense pipeline. Our results obtained from correlation analysis on all these metrics from different perspectives suggest that existing testing coverage criteria have limited correlation with the robustness (or the improvement of the robustness) of DNN models. Furthermore, we provide potential directions to improve DNN testing in general by correlation analysis of robustness metrics and other kinds of metrics. 

While our results are mostly negative, we believe it is important that future proposed testing criteria and methods undergo similar evaluation so as to provide evidence of their relevance. Our models, adversarial samples, and programs for calculating the metrics are publicly available and can be used as a benchmark for evaluating future research in this direction.